\documentclass{article}

\usepackage{PRIMEarxiv}
\usepackage{subcaption}
\usepackage[utf8]{inputenc} 
\usepackage[T1]{fontenc}    
\usepackage{hyperref}       
\usepackage{url}            
\usepackage{booktabs}       
\usepackage{amsfonts}       
\usepackage{nicefrac}       
\usepackage{microtype}      
\usepackage{lipsum}
\usepackage{fancyhdr}       
\usepackage{graphicx}       
\usepackage{amsmath}
\usepackage{multirow}
\usepackage{comment}

\graphicspath{{media/}}     

\title{XAI-Driven Diagnosis of Generalization Failure in State-Space Cerebrovascular Segmentation Models: A Case Study on Domain Shift Between RSNA and TopCoW Datasets
}

\author{
  Youssef Abuzeid \\
  Department of Electronics and Electrical Communications Engineering \\
  Cairo University \\
  \texttt{youssef.abuzeid01@eng-st.cu.edu.eg} \\
  \And
  Shimaa El-Bana \\
  Multimedia Interaction and Communication Lab \\
  Arab Academy for Science and Technology \\
  \texttt{shimaa.elbanaa@aiet.edu.eg} \\
  \And
  Ahmad Al-Kabbany \\
  Multimedia Interaction and Communication Lab \\
  Wearables, Biosensing, and Biosignal Processing Research lab \\
  Arab Academy for Science and Technology \\
  \texttt{alkabbany@ieee.org, alkabbany@aast.edu} \\
}

\begin{document}
\maketitle

\begin{abstract}
The clinical deployment of deep learning models in medical imaging is severely hindered by domain shift. This challenge, where a high-performing model fails catastrophically on external datasets, is a critical barrier to trustworthy AI. Addressing this requires moving beyond simple performance metrics toward deeper understanding, making Explainable AI (XAI) an essential diagnostic tool in medical image analysis. We present a rigorous, two-phase approach to diagnose the generalization failure of state-of-the-art State-Space Models (SSMs), specifically UMamaba, applied to cerebrovascular segmentation.  We first established a quantifiable domain gap between our Source (RSNA CTA Aneurysm) and Target (TopCoW Circle of Willis CT) datasets, noting significant differences in Z-resolution  and background noise.  The model's Dice score subsequently plummeted from 0.8604 (Source)  to 0.2902 (Target). In the second phase, which is our core contribution, we utilized Seg-XRes-CAM to diagnose the cause of this failure. We quantified the model's focus by measuring the overlap between its attention maps and the Ground Truth segmentations, and between its attention maps and its own Prediction Mask. Our analysis proves the model failed to generalize because its attention mechanism abandoned true anatomical features in the Target domain. Quantitative metrics confirm the model's focus shifted away from the Ground Truth vessels ($IoU\approx0.101$ at $0.3$ threshold)  while still aligning with its own wrong predictions ($IoU\approx0.282$ at $0.3$ threshold). This demonstrates the model learned spurious correlations, confirming XAI is a powerful diagnostic tool for identifying dataset bias in emerging architectures.
\end{abstract}


\section{Introduction}
\begin{figure}[h]
    \centering
    \includegraphics[width=0.7\linewidth]{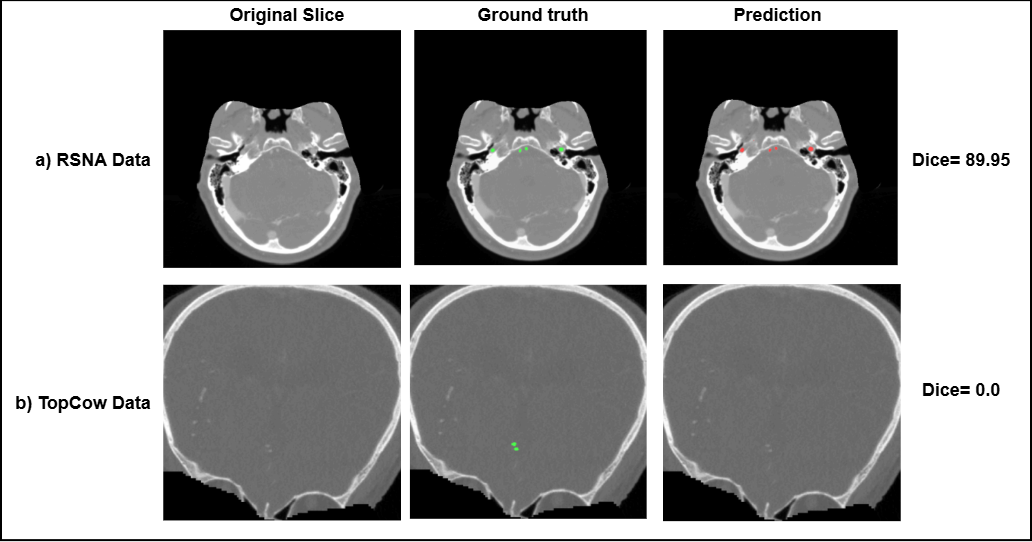}
    \caption{Inference results from a U-Mamba model trained on (a) RSNA, which performs well, but fails on (b) TopCoW, illustrating domain shift. The significant drop in Dice highlights the need for domain-robust methods.}
    \label{fig:domain_shift}
\end{figure}

The increasing availability of high-resolution, multimodal medical imaging data has driven a widespread reliance on deep learning (DL) models for automated diagnosis and quantification. This includes semantic segmentation tasks, such as identifying anatomical structures or pathologies \cite{du2024deep,zhang2024segment}. Segmentation models, including those used for cerebrovascular analysis \cite{bayram2025systematic}, have achieved performance parity with human experts on intra-domain datasets. This exceptional performance positions DL as a critical technology for scaling up clinical efficiency. However, the successful adoption and scaling of DL-based diagnosis hinge on trustworthiness \cite{shi2024survey}. A cornerstone of trustworthy AI in the high-stakes medical environment is ensuring that models remain robust and reliable when faced with real-world variability. This ambition is fundamentally challenged by the problem of domain shift \cite{mahmutoglu2025optimizing,yoon2024domain}. Domain shift is characterized by the catastrophic performance collapse of a high-performing model when deployed on external, unseen (Out-of-Distribution, OOD) data. This degradation is driven by subtle but systemic variations in factors like scanner manufacturers, imaging protocols, noise characteristics, and patient demographics that violate the model's training distribution. Medical image segmentation models often suffer from performance drops when applied to new domains (domain shift), as shown in Fig.~\ref{fig:domain_shift}.

Comparative evaluations in medical imaging often focus solely on quantifying what failed (performance drop) rather than diagnosing why the failure occurred. Consequently, it has become increasingly important to transition the focus from mere performance metrics to the internal decision-making logic of the models. In this context, simple aggregate metrics like the Dice Similarity Coefficient (Dice) and Intersection over Union (IoU) are insufficient for establishing the deep clinical trust required for deployment. To bridge this gap, Explainable AI (XAI) has emerged as the critical diagnostic tool needed to break the "black-box" nature of advanced models and gain the deeper understanding required for reliable deployment in safety-critical domains \cite{bhati2024survey}. Explainable AI (XAI) methods, primarily based on feature attribution, aim to highlight the regions of the input image that most influenced the model's prediction \cite{houssein2025explainable}. These techniques range from classification-oriented methods like Grad-CAM to specialized variants for dense prediction tasks, such as Seg-XRes-CAM \cite{hasany2024the,10208849}. Unlike traditional Grad-CAM, which explains image-level classification, Seg-XRes-CAM is specifically designed to address the challenges of segmentation by preserving critical spatial resolution for pixel-wise explanations.

One significant domain in medical segmentation that has highly benefited from the power of deep learning is cerebrovascular segmentation \cite{hu2024deep,yagis2024deep,ince2025deep}. This critical task, which involves isolating vessel structures from multimodal datasets in the literature, inherently suffers from domain shift due to variations in imaging protocols (e.g., CTA vs. MRA) and clinical context. The nature of cerebrovascular segmentation presents numerous complexities, including the presence of fine, thread-like structures that require high spatial resolution, extensive branching complexity, and variable contrast agents. These inherent challenges have continually stimulated the reliance on advanced, high-capacity architectures to capture the required long-range dependencies and fine-grained details. Recently, State-Space Models (SSMs) have emerged as the current architectural frontier, offering an efficient alternative to the computationally heavy Attention mechanism of Transformers \cite{bansal2024comprehensive}. Architectures like U-Mamba, which integrate the selective state-space modeling into an encoder-decoder structure, have demonstrated high intra-domain performance by effectively modeling long-range dependencies \cite{xu2025mambavesselnet++,chen2024mambavesselnet,elbana2025efficiency}. However, as our study demonstrates, despite achieving high performance on their training distributions, these advanced SSMs remain severely vulnerable to catastrophic generalization failure when faced with external, unseen clinical data, highlighting a critical flaw in their feature robustness.

Inspired by this vulnerability, our work is motivated by a significant gap in the current literature. While extensive research focuses on solution-oriented methods like Domain Adaptation (DA) \cite{wang2024unsupervised}, feature disentanglement, or architectural fixes to mitigate domain shift, there remains a critical deficiency in providing a mechanistic, XAI-driven diagnosis of why the failure occurs in OOD data. Addressing this, we propose a novel two-phase diagnostic approach. The first phase aims to rigorously quantify the domain gap and performance degradation by comparing dataset statistics—including Z-resolution, background noise, and vessel morphology—between the Source (RSNA) \cite{rsna_ica_dataset_2025} and Target (TopCoW) domains \cite{yang2025benchmarking}, representing two recent datasets in the literature. The second phase, which is the core of our diagnosis, employs XAI to break the model's black box and visually and quantitatively diagnose the root cause of the feature failure. To achieve this rigor, we adopt a recent XAI evaluation framework that involves comparing the model's attention map against both the Ground Truth (GT) mask and the model's Prediction Mask (PM), enabling us to pinpoint the specific mechanisms of spurious correlation.

Through our rigorous two-phase diagnostic approach, we conclusively identified the mechanism behind the observed failure. The results of the first phase established a substantial, quantified domain gap (35.8\% Z-resolution difference and a 3.4x background noise increase) that led to the catastrophic performance collapse of the U-Mamba model, with the Dice score plummeting from on the Source domain to on the Target domain. The second phase, using Seg-XRes-CAM and our quantitative framework, provided the decisive evidence: on the Target domain, the model's attention entirely abandoned true anatomical features, evidenced by a low XAI-GT ( at threshold), while simultaneously remaining aligned with its own erroneous predictions (XAI-PM at threshold). This pattern is the definitive signature of spurious correlation: the model confidently overfits to non-generalizable, domain-specific artifacts. Our contributions are summarized as follows:
\begin{itemize}
    \item A two-phase diagnostic framework is introduced, combining quantitative domain gap analysis with segmentation-specific explainability to investigate generalization failure in U-Mamba.

    \item A quantified assessment of the RSNA $\rightarrow$ TopCoW domain shift is provided, revealing key differences in Z-resolution, background noise, and vessel morphology that correspond to a severe performance collapse.
    
    \item An explainability evaluation strategy is presented using Seg-XRes-CAM with multi-layer aggregation and two alignment metrics—XAI-GT and XAI-PM—to measure how model attention relates to true anatomy and predicted structures.
    
    \item Evidence of spurious correlations is demonstrated through low attention--ground-truth alignment and higher attention--prediction alignment on the target domain, indicating reliance on non-anatomical features.
    
    \item A slice-level diagnostic taxonomy is proposed to reveal distinct failure modes and expose systematic breakdowns not captured by standard segmentation metrics.
\end{itemize}

The remainder of this paper is structured as follows: Section 2 reviews the related literature on domain shift and XAI; Section 3 quantifies the domain gap; Section 4 details the segmentation model, the Seg-XRes-CAM methodology, and the XAI evaluation framework; Section 5 presents the quantitative and qualitative diagnostic results; and Section 6 concludes the article, offering insights for future domain generalization research.

\section{Related Work}
\label{sec:related_work}

The deployment of deep learning models for vascular segmentation is frequently hindered by domain shift, a phenomenon that arises from variations in scanner manufacturers, imaging protocols, and patient demographics \cite{atwany2025angiodginterpretablechannelinformedfeaturemodulated, gao2023bayesegbayesianmodelingmedical}. Although convolutional neural networks (CNNs) and recent state-space models have achieved high performance on intra-domain tasks, they often fail to generalize to unseen target domains (Out-of-Distribution, OOD data) due to overfitting to source-domain statistics \cite{atwany2025angiodginterpretablechannelinformedfeaturemodulated, gao2023bayesegbayesianmodelingmedical}. 

To mitigate this, recent literature has explored Domain Adaptation (DA) and Domain Generalization (DG). In the context of cerebrovascular segmentation, Galati et al. proposed a semi-supervised DA framework that employs feature disentanglement to translate images between modalities (e.g., MRA to CTA or MRV) while preserving vessel geometry via a label-synthesis mechanism \cite{Galati_2025}. However, DA methods often require access to target domain data or computationally expensive retraining, making Single-Source Domain Generalization (SDG) a more practical clinical alternative \cite{atwany2025angiodginterpretablechannelinformedfeaturemodulated, hu2023devilchannelscontrastivesingle}.

A prevailing hypothesis in recent SDG literature is that domain-specific style information which hinders generalization is encoded in specific feature channels, particularly within shallow layers. Hu et al. identified that "the devil is in the channels" (C2SDG), observing that certain shallow channels are highly sensitive to style variations while others encode structure \cite{hu2023devilchannelscontrastivesingle}. They proposed a contrastive learning framework to explicitly disentangle these features, discarding style components during inference \cite{hu2023devilchannelscontrastivesingle}. 

Similarly, Atwany et al. introduced \textit{AngioDG} for coronary vessel segmentation, which employs channel whitening to decorrelate features and a Weighted Channel Attention (WCA) module to amplify domain-invariant channels while suppressing those prone to overfitting background artifacts \cite{atwany2025angiodginterpretablechannelinformedfeaturemodulated}. In the domain of brain vessel segmentation, Galati et al. utilized path length regularization within a StyleGAN-based architecture to disentangle vessel-specific properties (e.g., intensity, texture) from volume-related properties, allowing label-preserving translations across large domain gaps \cite{Galati_2025}.

Beyond channel manipulation, explicit statistical modeling has been explored to separate anatomical structure from image appearance. \textit{BayeSeg} employs a Bayesian framework to decompose images into a spatially correlated shape variable (domain-invariant) and a spatially variable appearance variable, enforcing hierarchical priors to improve interpretability and generalization \cite{gao2023bayesegbayesianmodelingmedical}. 

Furthermore, to mitigate the sensitivity of pixel-wise classification to local intensity shifts, Zhao et al. proposed \textit{ISAC}, which reformulates vascular segmentation as a mask completion task \cite{10.1007/978-3-032-04981-0_29}. By reconstructing vessels from sparse supports and using an Uncertainty-guided Patch selection module (UPS), ISAC prioritizes the structural continuity of the vascular tree over texture, enhancing robustness against domain shifts in zero-shot scenarios \cite{10.1007/978-3-032-04981-0_29}.

Although the aforementioned methods propose architectural solutions to mitigate domain shift, there remains a critical gap in rigorously diagnosing the \textit{underlying causes} of generalization failure in state-of-the-art architectures, particularly with regard to spurious correlations. Existing work largely focus on performance metrics (e.g., Dice, Hausdorff Distance) or solution-oriented disentanglement, often treating the model as a "black box" regarding why specific features trigger failure in OOD data.

Unlike previous works that assume the dependence of features based on performance drops, our study employs Explainable AI (XAI) as a diagnostic tool. We build upon the need for interpretability highlighted in medical imaging literature to visualize spurious correlations such as model reliance on scanner artifacts or background noise that drive generalization failure. This work creates a bridge between quantitative performance gaps and visual feature diagnosis, with the aim of providing a mechanistic understanding of domain shift in cerebrovascular segmentation.

\section{Dataset Descriptions }
\label{sec:datasets}
This study employs two publicly available multi-modal datasets to develop and evaluate the proposed framework: the RSNA Intracranial Aneurysm Detection dataset and the TopCoW dataset. Both datasets contain volumetric brain imaging data across multiple modalities, including computed tomography (CT), magnetic resonance angiography (MRA), and magnetic resonance imaging (MRI).

\subsection{Source Domain: RSNA CTA Aneurysm Dataset}
The Source domain consists of Computed Tomography Angiography (CTA) volumes from the RSNA Intracranial Hemorrhage Detection challenge \footnote{\url{https://www.kaggle.com/competitions/rsna-intracranial-aneurysm-detection/data}}, containing cerebrovascular annotations including aneurysms. The dataset exhibits an in-plane resolution of approximately 0.457 mm and a slice thickness of 1.114 mm. Vessel annotations span a wide range of diameters, from small perforating arteries to large aneurysmal structures, with diameters ranging from 0.5 mm to 26.7 mm (mean: 2.43 mm, std: 1.58 mm).

\subsection{Target Domain: TopCoW Circle of Willis CT Dataset}
The Target domain comprises CT volumes from the TopCoW challenge \footnote{\url{https://topcow24.grand-challenge.org/data/}}, focusing specifically on Circle of Willis segmentation. The dataset features similar in-plane resolution (0.452 mm) but significantly finer slice thickness (0.715 mm). Vessel annotations are more anatomically constrained to the Circle of Willis structure, with diameters ranging from 0.6 mm to 5.9 mm (mean: 2.31 mm, std: 0.77 mm).

\section{Domain Shift Quantification}
\label{sec:domain_quantification}

To quantify the domain gap between the Source (RSNA) and Target (TopCoW) datasets, we provided a systematic analysis focusing on four key aspects that can impact segmentation performance.: intensity distribution, spatial resolution, noise characteristics, and vessel morphology. This analysis provides a clear characterization of differences that may affect model generalization across the two domains.

\paragraph{Intensity Analysis}
Within the standard soft tissue window (0--80 HU), both datasets exhibit remarkably similar intensity characteristics. Table~\ref{tab:intensity} summarizes the intensity statistics, revealing nearly identical mean intensities (41.71 HU vs. 42.11 HU) and standard deviations (16.99 HU vs. 17.52 HU). This similarity indicates that the domain shift is not primarily driven by differences in tissue contrast or brightness within the clinically relevant intensity range.

\begin{table}[h]
\caption{Intensity Distribution Comparison (0--80 HU Window)}
\centering
\begin{tabular}{lcc}
\toprule
\textbf{Metric} & \textbf{Source (RSNA)} & \textbf{Target (TopCoW)} \\
\midrule
Mean Intensity (HU) & 41.71 & 42.11 \\
Std Intensity (HU) & 16.99 & 17.52 \\
\bottomrule
\end{tabular}
\label{tab:intensity}
\end{table}

\paragraph{Resolution Analysis}
A critical difference emerges in the Z-axis resolution. As shown in Table~\ref{tab:resolution}, while in-plane resolution remains comparable across domains ($\sim$0.45 mm), the Target dataset exhibits 36\% finer slice thickness (0.715 mm vs. 1.114 mm). This disparity introduces significant differences in the volumetric representation of vessel structures, particularly affecting the continuity and smoothness of vessels along the axial direction.

\begin{table}[h]
\caption{Spatial Resolution Comparison}
\centering
\begin{tabular}{lccc}
\toprule
\textbf{Axis} & \textbf{Source (RSNA)} & \textbf{Target (TopCoW)} & \textbf{Difference} \\
\midrule
X Resolution (mm) & 0.457 & 0.452 & -1.1\% \\
Y Resolution (mm) & 0.457 & 0.452 & -1.1\% \\
Z Resolution (mm) & 1.114 & 0.715 & \textbf{-35.8\%} \\
\bottomrule
\end{tabular}
\label{tab:resolution}
\end{table}

\paragraph{Noise Analysis}
Background noise analysis reveals a substantial difference between domains. The Target dataset exhibits a background noise standard deviation of 7.96 HU compared to 2.37 HU in the Source domain---a 3.4$\times$ increase (Table~\ref{tab:noise}). This elevated noise level poses significant challenges for segmentation models, particularly for detecting smaller or lower-contrast vessel structures that may become obscured by noise artifacts.

\begin{table}[h]
\caption{Background Noise Comparison}
\centering
\begin{tabular}{lcc}
\toprule
\textbf{Metric} & \textbf{Source (RSNA)} & \textbf{Target (TopCoW)} \\
\midrule
Noise Std (HU) & 2.37 & 7.96 \\
Noise Ratio & -- & \textbf{3.4$\times$} \\
\bottomrule
\end{tabular}
\label{tab:noise}
\end{table}

\paragraph{Vessel Morphology Analysis}
Vessel diameter analysis reveals distinct morphological differences between the datasets (Table~\ref{tab:vessel_diameter}). While mean diameters are comparable (2.43 mm vs. 2.31 mm), the Source dataset exhibits substantially greater variability (std: 1.58 mm vs. 0.77 mm) and contains extreme outliers with maximum diameters exceeding 26 mm, likely corresponding to aneurysmal structures. In contrast, the Target dataset, focused on normal Circle of Willis anatomy, demonstrates a more constrained diameter distribution (max: 5.93 mm). This difference in vessel morphology reflects the distinct clinical contexts of the two datasets.

\begin{table}[h]
\caption{Vessel Diameter Distribution Comparison}
\centering
\begin{tabular}{lcc}
\toprule
\textbf{Metric} & \textbf{Source (RSNA)} & \textbf{Target (TopCoW)} \\
\midrule
Mean Diameter (mm) & 2.43 & 2.31 \\
Median Diameter (mm) & 2.07 & 2.25 \\
Std Deviation (mm) & \textbf{1.58} & 0.77 \\
Min Diameter (mm) & 0.50 & 0.60 \\
Max Diameter (mm) & \textbf{26.74} & 5.93 \\
99th Percentile (mm) & 9.39 & 4.46 \\
\bottomrule
\end{tabular}
\label{tab:vessel_diameter}
\end{table}

Overall, the domain shift is driven by three main factors. First, the 36\% reduction in slice thickness in the Target dataset alters vessel continuity and volumetric representation. Second, the 3.4$\times$ higher background noise can obscure vessel boundaries and complicate segmentation. Third, vessel morphology differences—reduced diameter variability and absence of large aneurysms in the Target dataset—further distinguish the domains. These quantifiable differences define a measurable domain gap and motivate our subsequent explainability-driven analysis of model generalization failure.

\section{Proposed Method}
\label{sec:methods}
\begin{figure*}[t]
    \centering
    \includegraphics[width=\textwidth]{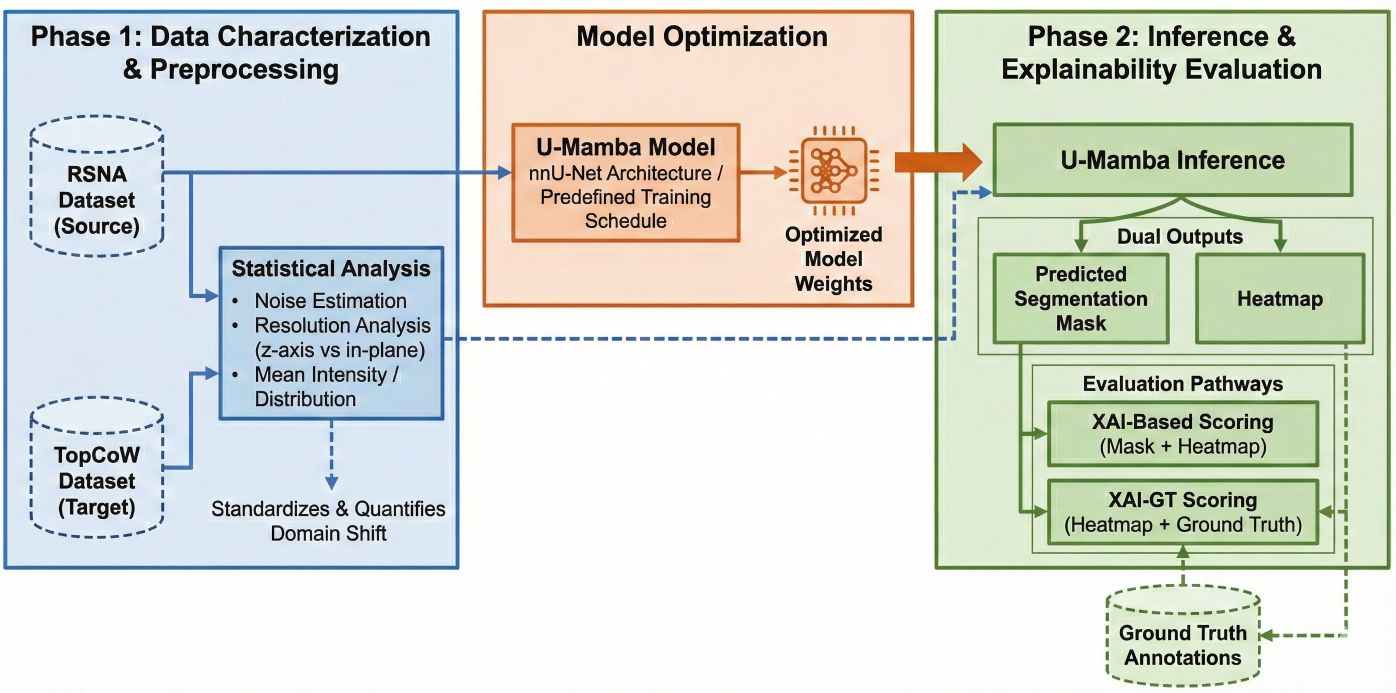} 
    \caption{Overview of the proposed two-phase diagnostic framework. \textbf{Phase 1} quantifies the domain shift through statistical analysis of noise, resolution, and intensity distributions between RSNA (Source) and TopCoW (Target). The \textbf{Model Optimization} stage trains the U-Mamba architecture on the source domain. \textbf{Phase 2} employs Seg-XRes-CAM to generate dual outputs (segmentation masks and heatmaps), evaluating generalization failure via two alignment pathways: XAI-GT (Attention vs. Ground Truth) and XAI-Based Scoring (Attention vs. Prediction).}
    \label{fig:framework}
\end{figure*}
To rigorously diagnose the generalization failure of state-space models under domain shift, we propose a multi-stage framework, as illustrated in Fig.~\ref{fig:framework}. The pipeline is structured into two distinct phases to bridge the gap between statistical data differences and internal feature representations.

\textbf{Phase 1 (Data Characterization \& Preprocessing)} focuses on quantifying the domain gap prior to inference. As detailed in Section~\ref{sec:domain_quantification}, we perform a comparative statistical analysis of the Source (RSNA) and Target (TopCoW) datasets, explicitly measuring shifts in noise levels, spatial resolution (specifically Z-axis anisotropy), and intensity distributions. This phase establishes the "ground truth" of the domain shift against which model performance is evaluated.

\textbf{Phase 2 (Inference \& Explainability Evaluation)} constitutes the core diagnostic stage. The U-Mamba model, optimized on the Source domain, generates dual outputs for every input volume: a predicted segmentation mask and a corresponding high-resolution attention heatmap via Seg-XRes-CAM. These outputs feed into two parallel evaluation pathways:
\begin{itemize}
    \item \textbf{XAI-GT Scoring:} Measures the alignment between the heatmap and the Ground Truth annotations to assess anatomical faithfulness.
    \item \textbf{XAI-Based Scoring (XAI-PM):} Measures the alignment between the heatmap and the model's own predictions to assess internal consistency.
\end{itemize}
By correlating the results from Phase 1 and Phase 2, we determine whether performance drops are due to simple statistical mismatches or deeper spurious correlations.
\subsection{Cerebrovascular Segmentation Model}

We adopt U-Mamba \cite{ma2024umambaenhancinglongrangedependency} as our segmentation backbone. U-Mamba is a state-space model (SSM)–based architecture designed to overcome the limitations of conventional CNNs and Transformers. While CNNs struggle to capture long-range dependencies due to their local receptive fields, Transformer-based models incur quadratic complexity with respect to feature resolution. U-Mamba addresses both issues by integrating selective state-space modeling into a U-Net–style encoder–decoder framework, enabling efficient global context modeling with linear computational cost.

The network consists of a symmetric encoder–decoder connected by skip connections. Its core component is the U-Mamba Block, which fuses local and global feature extraction. Each block includes two Residual units—each composed of a convolutional layer, Instance Normalization, and Leaky ReLU—followed by a Mamba block. The Mamba block is responsible for long-range dependency modeling: input features are flattened into a sequence of shape $(B, L, C)$, where $L$ corresponds to the spatial dimension (e.g., $H \times W$ for 2D). The block applies Layer Normalization, linear projections, a 1D convolution, and an SSM layer, merged through a Hadamard product across parallel branches.

In the decoder, standard Residual blocks and transposed convolutions progressively restore spatial resolution and refine local anatomical details. The model is trained on the Source domain (RSNA CTA Aneurysm) and achieves a Dice score of 0.8604, establishing a strong baseline for evaluating cross-domain generalization.

\subsection{Explainability Method}
\label{sec:segxrescam}

To investigate the causes of cross-domain generalization failure, we adopt Seg-XRes-CAM \cite{10208849}, a gradient-based explainability method tailored for semantic segmentation. 

\subsubsection{Justification of Method Selection}

Standard XAI methods for classification are ill-suited for dense prediction tasks. We performed a comparative evaluation of three explainability configurations to select the optimal tool for our diagnostic framework. We evaluated \textbf{Seg-Grad-CAM} \cite{Vinogradova_2020}, and two variants of \textbf{Seg-XRes-CAM} \cite{10208849} with differing gradient pooling window sizes: Max Pool 1 ($1\times1$ kernel) and Max Pool 2 ($2\times2$ kernel).

\begin{figure*}[t]
    \centering
    \includegraphics[width=\linewidth]{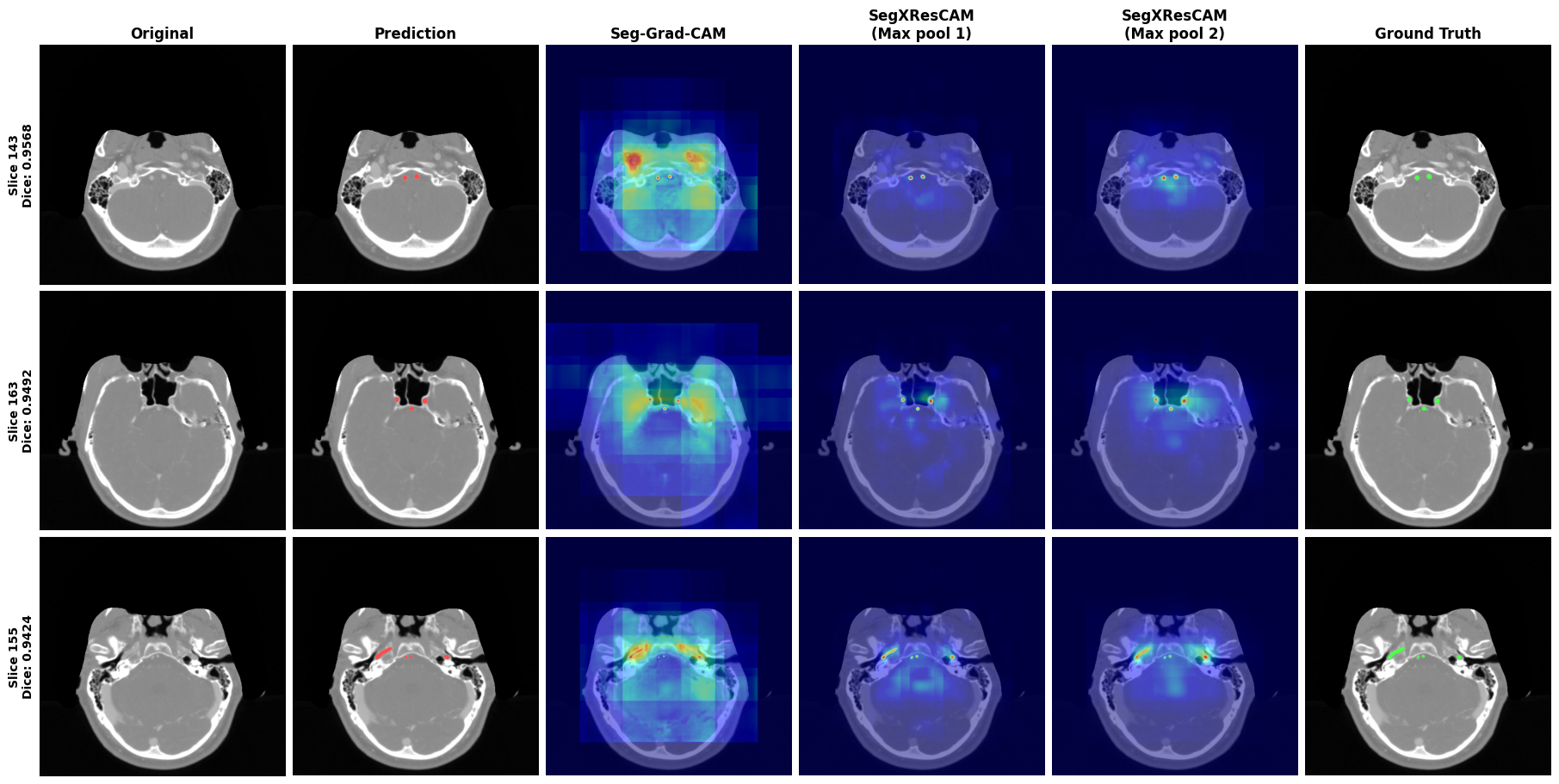} 
    \caption{Methodological validation on the Source domain. \textbf{Seg-Grad-CAM} generates diffuse, low-resolution heatmaps. \textbf{SegXResCAM (Max Pool 2)} improves localization but over-smoothes fine details. \textbf{SegXResCAM (Max Pool 1)} provides the highest spatial fidelity, accurately capturing the fine vascular tree required for diagnosing domain shift.}
    \label{fig:xai_method_comparison}
\end{figure*}

Qualitatively, the differences are distinct (Figure~\ref{fig:xai_method_comparison}). Seg-Grad-CAM produces coarse, blocky heatmaps due to the upsampling of low-resolution bottleneck features. Seg-XRes-CAM (Max Pool 2) offers better localization but suffers from over-smoothing, dilating the attention regions beyond the actual vessel boundaries. In contrast, Seg-XRes-CAM (Max Pool 1) provides the highest spatial fidelity, sharply delineating the fine, thread-like structures of the Circle of Willis.

The quantitative results, summarized in Table~\ref{tab:xai_method_comparison}, confirm these visual observations. Seg-Grad-CAM achieves a negligible IoU of 0.0102 on the Source domain. While Max Pool 2 improves this to 0.1863, it still lags significantly behind \textbf{Max Pool 1}, which achieves the highest alignment (IoU 0.4671) and F1 Score (0.6368). Consequently, Seg-XRes-CAM (Max Pool 1) was selected as the primary diagnostic tool for this study.

\begin{table}[h]
\caption{Comparative evaluation of Explainability Methods at threshold $\tau=0.3$. \textbf{SegXResCAM (Max Pool 1)} yields the highest alignment (IoU/F1) on the Source domain, minimizing the over-smoothing observed in Max Pool 2 and the coarse resolution of Seg-Grad-CAM.}
\centering
\begin{tabular}{llcccc}
\toprule
\textbf{Method} & \textbf{Domain} & \textbf{IoU (XAI-GT)} & \textbf{F1 Score} & \textbf{Precision} & \textbf{Recall} \\
\midrule
\multirow{2}{*}{Seg-Grad-CAM} 
  & Source & 0.0102 & 0.0201 & 0.0102 & 0.7193 \\
  & Target & 0.0055 & 0.0110 & 0.0056 & 0.5427 \\
\midrule
\multirow{2}{*}{SegXResCAM (Max Pool 2)} 
  & Source & 0.1863 & 0.3141 & 0.1933 & \textbf{0.8381} \\
  & Target & 0.0641 & 0.1197 & 0.0682 & 0.5266 \\
\midrule
\multirow{2}{*}{\textbf{SegXResCAM (Max Pool 1)}} 
  & Source & \textbf{0.4671} & \textbf{0.6368} & \textbf{0.6665} & 0.6097 \\
  & Target & 0.1018 & 0.1832 & 0.1373 & 0.3222 \\
\bottomrule
\end{tabular}

\label{tab:xai_method_comparison}
\end{table}

\subsubsection{Implementation Details}
\paragraph{Multi-Layer Mean Aggregation:}
Hasany et al. \cite{hasany2024the} demonstrated that relying on a single layer in segmentation models can produce incomplete explanations. Following their recommendations, we extract saliency maps from multiple layers and aggregate them via mean pooling:

\begin{equation}
S_{\text{final}} = \frac{1}{|L|} \sum_{l \in L} S_l
\end{equation}

where $L$ represents our selected layers: \texttt{encoder.4}, \texttt{encoder.5}, and \texttt{decoder.0} through \texttt{decoder.4}.

\paragraph{Target Selection:}
The gradient target for Seg-XRes-CAM is set to the model's own prediction mask rather than the ground truth. This design choice is critical: we aim to explain \textit{what the model actually predicts} and \textit{why}, enabling the detection of spurious correlations when the model's attention diverges from true anatomical features.
\subsection{Explainability Evaluation Framework}
\label{sec:xai_evaluation}

To quantitatively assess the quality and faithfulness of our saliency maps, we adopt the evaluation framework proposed by Hammoud et al. \cite{hammoud2025quantitativeevaluationframeworkexplainable}, which provides a rigorous methodology for evaluating explainable AI methods in semantic segmentation. This framework enables systematic comparison of attention alignment with both ground truth anatomy and model predictions.

\paragraph{Attention–Ground Truth Overlap (XAI-GT IoU):}
This metric quantifies how well the model’s attention corresponds to the true anatomical structures. We compute the intersection-over-union between the binarized attention map $A_\tau$ (thresholded at level $\tau$) and the ground truth mask $M_{GT}$:
\begin{equation}
\text{IoU}_{\text{XAI-GT}} = \frac{|A_\tau \cap M_{GT}|}{|A_\tau \cup M_{GT}|}.
\end{equation}
A high XAI-GT IoU indicates that the saliency map correctly highlights vessel regions, reflecting anatomically faithful attention patterns. Conversely, a low score suggests that the model attends to non-vascular regions or background structures, providing direct evidence of misalignment between the learned features and clinically relevant anatomy.

\paragraph{Attention–Prediction Overlap (XAI-PM IoU):}
This metric evaluates the internal consistency of the model by measuring how well its attention aligns with its own predicted segmentation mask $M_{pred}$. It is computed as the intersection-over-union between the binarized attention map $A_\tau$ and $M_{pred}$:
\begin{equation}
\text{IoU}_{\text{XAI-PM}} = \frac{|A_\tau \cap M_{pred}|}{|A_\tau \cup M_{pred}|}.
\end{equation}
A high XAI-PM IoU indicates that the model’s predictions are supported by the regions it considers important, suggesting coherent internal reasoning. In contrast, a low score reveals a mismatch between attention and output, implying either diffuse, uninformative saliency or reliance on features that do not directly contribute to the final prediction.

Furthermore, we evaluate the saliency maps using precision, recall, and F1 score following the framework of \cite{hammoud2025quantitativeevaluationframeworkexplainable}. Precision quantifies the proportion of the highlighted attention that corresponds to true anatomical regions:
\begin{equation}
\text{Precision} = \frac{|A_\tau \cap M_{GT}|}{|A_\tau|}.
\end{equation}
Recall measures how much of the ground truth anatomy is successfully captured by the attention map:
\begin{equation}
\text{Recall} = \frac{|A_\tau \cap M_{GT}|}{|M_{GT}|}.
\end{equation}
The F1 score provides a balanced summary of these two aspects:
\begin{equation}
\text{F1} = \frac{2 \times \text{Precision} \times \text{Recall}}{\text{Precision} + \text{Recall}}.
\end{equation}
Together, these metrics, which are summarized in  Table \ref{tab:eval-metrics}, offer complementary insights into the relevance and completeness of the model’s attention, enabling a more nuanced assessment beyond IoU-based measures.

\paragraph{Diagnostic Logic}
The relationship between XAI-GT and XAI-PM metrics reveals the nature of model failure:
\begin{itemize}
    \item {High XAI-GT, High XAI-PM}: Model attends to correct features and predicts accurately.
    \item {Low XAI-GT, Low XAI-PM}: Model attention is diffuse or noisy.
    \item {Low XAI-GT, Moderate XAI-PM}: Model confidently attends to wrong features evidence of \textit{spurious correlation}.
\end{itemize}
The third scenario is particularly diagnostic: the model maintains internal consistency (attention aligns with predictions) but has abandoned true anatomical features, indicating overfitting to domain-specific artifacts.

\paragraph{Threshold Selection}
We evaluated thresholds $\tau \in \{0.2, 0.3, 0.4\}$ for binarizing saliency maps. Threshold $\tau = 0.3$ yielded the best F1 score balance between precision and recall of attention regions, and is used for all reported metrics.

\begin{table}[t]
    \centering
    \caption{Summary of Evaluation Metrics}
    \label{tab:eval-metrics}
    \begin{tabular}{llp{0.3\textwidth}}
        \hline
        \textbf{Category} & \textbf{Metric} & \textbf{Purpose} \\
        \hline
        \multirow{2}{*}{Segmentation}
            & Dice (DSC) & Volumetric overlap \\ 
            & IoU        & Alternative overlap measure \\
        \hline
        \multirow{5}{*}{Explainability}
            & XAI-GT IoU & Attention-anatomy alignment \\
            & XAI-PM IoU & Attention-prediction alignment \\
            & Precision  & Relevance of attention \\
            & Recall     & Coverage of anatomy \\
            & F1 Score   & Balanced measure \\
        \hline
    \end{tabular}
\end{table}

\section{Results and Discussion}
\subsection{Cross-Domain Segmentation Performance}
\label{sec:results_segmentation}

Establishing a baseline for the generalization gap involved evaluating the U-Mamba model, trained on the Source domain (RSNA CTA Aneurysm), on both the Source and Target (TopCoW) datasets. The resulting volumetric Dice scores are presented in Table~\ref{tab:segmentation_performance}.

\begin{table}[h]
\caption{Cross-Domain Segmentation Performance}
\centering
\begin{tabular}{lcc}
\toprule
\textbf{Domain} & \textbf{Dice} & \textbf{Performance Drop} \\
\midrule
Source (RSNA) & 0.860 & -- \\
Target (TopCoW) & 0.2902 & \textbf{-66.3\%} \\
\bottomrule
\end{tabular}
\label{tab:segmentation_performance}
\end{table}

The Dice score decreases sharply from 0.860 on the Source domain to 0.2902 on the Target domain, corresponding to a 66.3\% performance reduction. This pronounced degradation clearly reflects the presence of substantial domain shift and demonstrates the model's limited ability to generalize to unseen clinical distributions.

A deeper understanding of this degradation was obtained by examining slice-level Dice scores and grouping slices according to predefined performance thresholds. Table~\ref{tab:slice_distribution} summarizes the distribution across these categories for both domains.

\begin{table}[h]
\caption{Slice-Level Performance Distribution}
\centering
\begin{tabular}{llrr}
\toprule
\textbf{Category} & \textbf{Dice Threshold} & \textbf{Source (\%)} & \textbf{Target (\%)} \\
\midrule
Perfect & $> 0.85$ & 18.4\% & 0.0\% \\
Good & $0.65 < \text{Dice} \leq 0.85$ & 6.0\% & 0.9\% \\
Bad & $0.4 < \text{Dice} \leq 0.65$ & 3.2\% & 5.6\% \\
Worst & $\leq 0.4$ & 1.8\% & 43.8\% \\
True Negative & No vessels (correct) & 70.5\% & 49.7\% \\
\bottomrule
\end{tabular}
\label{tab:slice_distribution}
\end{table}

On the Source dataset, 24.4\% of slices lie within the good-to-perfect range (Dice $> 0.65$), and only 1.8\% fall into the worst-performing category. In contrast, the Target dataset shows only 0.9\% of slices achieving good performance, while 43.8\% exhibit severe failure (Dice $\leq 0.4$). This marked shift in distribution indicates that the domain shift introduces widespread, systematic degradation rather than isolated prediction errors.
\subsection{Explainability Analysis}
\label{sec:results_xai}

Following the identification of the performance gap, Seg-XRes-CAM was employed to investigate the underlying mechanisms contributing to the model's generalization failure. The analysis focused on two quantitative alignment metrics: XAI-GT IoU, which measures the correspondence between the attention map and ground-truth vessels, and XAI-PM IoU, which quantifies the alignment between attention and the model’s predicted masks. Each metric was evaluated across multiple thresholds to assess robustness.

Table~\ref{tab:xai-gt-alignment} summarizes the XAI-GT results. On the Source domain, the attention maps exhibit substantial anatomical alignment; for instance, at a threshold of $\tau = 0.3$, the XAI-GT IoU reaches 0.4671. In contrast, the Target domain shows a marked reduction to 0.1018 at the same threshold, corresponding to a 78.2\% decline. This pronounced reduction indicates that model attention largely diverges from true vessel structures under domain shift.

\begin{table}[t]
    \centering
    \caption{XAI-Ground Truth Alignment (XAI $\cap$ GT)}
    \label{tab:xai-gt-alignment}
    \begin{tabular}{llcccc}
        \hline
        \textbf{Threshold} & \textbf{Domain} & \textbf{IoU} & \textbf{F1} & \textbf{Precision} & \textbf{Recall} \\
        \hline
        \multirow{2}{*}{0.2} 
            & Source & 0.2251 & 0.3675 & 0.2380 & 0.8061 \\
            & Target & 0.0545 & 0.1025 & 0.0571 & 0.5836 \\
        \hline
        \multirow{2}{*}{0.3} 
            & Source & 0.4671 & 0.6368 & 0.6655 & 0.6097 \\
            & Target & \textbf{0.1018} & 0.1832 & 0.1373 & 0.3222 \\
        \hline
        \multirow{2}{*}{0.4} 
            & Source & 0.4143 & 0.5858 & 0.9406 & 0.4254 \\
            & Target & 0.1002 & 0.1803 & 0.2227 & 0.1581 \\
        \hline
    \end{tabular}
\end{table}

The XAI-PM results in Table~\ref{tab:xai-pm-alignment} show a noticeably higher alignment between attention and predictions on the Target domain compared to XAI-GT. At $\tau = 0.3$, the Target XAI-PM IoU reaches 0.2823—almost three times the XAI-GT value. This indicates that the model remains internally consistent, but its attention is aligned with incorrect, non-anatomical predictions.

\begin{table}[t]
    \centering
    \caption{XAI-Prediction Mask Alignment (XAI $\cap$ PM)}
    \label{tab:xai-pm-alignment}
    \begin{tabular}{llcccc}
        \hline
        \textbf{Threshold} & \textbf{Domain} & \textbf{IoU} & \textbf{F1} & \textbf{Precision} & \textbf{Recall} \\
        \hline
        \multirow{2}{*}{0.2}
            & Source & 0.2258 & 0.3684 & 0.2328 & 0.8819 \\
            & Target & 0.0913 & 0.1647 & 0.0926 & 0.8982 \\
        \hline
        \multirow{2}{*}{0.3}
            & Source & 0.5220 & 0.6859 & 0.6782 & 0.6938 \\
            & Target & \textbf{0.2823} & 0.4249 & 0.3358 & 0.7161 \\
        \hline
        \multirow{2}{*}{0.4}
            & Source & 0.4781 & 0.6469 & 0.9630 & 0.4871 \\
            & Target & 0.3489 & 0.5029 & 0.6774 & 0.4767 \\
        \hline
    \end{tabular}
\end{table}

The divergence between XAI-GT and XAI-PM is further illustrated in Figure~\ref{fig:xai_comparison_metrics}. On the Source domain, the two metrics remain closely aligned (gap = 0.0549), indicating that attention supports both anatomical accuracy and predictive consistency. Under domain shift, however, the gap widens substantially to 0.1805. This characteristic pattern—low XAI-GT combined with moderate XAI-PM—constitutes clear evidence of spurious correlation, suggesting the model relies on domain-specific artifacts rather than anatomical cues.

\begin{figure}[h]
\centering
\begin{tabular}{lcc}
\toprule
\textbf{Metric} & \textbf{Source} & \textbf{Target} \\
\midrule
XAI-GT IoU & 0.4671 & 0.1018 \\
XAI-PM IoU & 0.5220 & 0.2823 \\
\midrule
\textbf{Gap (XAI-PM $-$ XAI-GT)} & 0.0549 & \textbf{0.1805} \\
\bottomrule
\end{tabular}
\caption{Comparison of XAI alignment metrics at threshold 0.3. The widening gap between XAI-PM and XAI-GT on the Target domain indicates a shift in attention towards non-anatomical features.}
\label{fig:xai_comparison_metrics}
\end{figure}

A joint evaluation of segmentation Dice scores and explainability F1 scores provides additional insight into failure modes. Table~\ref{tab:joint-slice-analysis} presents the distribution of slices across four diagnostic categories. On the Source domain, 75.0\% of vessel-containing slices exhibit both correct segmentation and meaningful attention. Conversely, the Target domain shows a collapse in this correspondence: only 2.6\% of slices fall into the same favorable category, while 92.9\% fall into the worst-performing category, characterized by both inaccurate segmentation and anatomically irrelevant attention.

\begin{table}[t]
    \centering
    \caption{Joint Slice-Level Analysis: Segmentation vs. Explainability Quality}
    \label{tab:joint-slice-analysis}
    \begin{tabular}{llrcrr}
        \hline
        \textbf{Domain} & \textbf{Category} & \textbf{Count} & \textbf{\%} &
        \textbf{Mean Dice} & \textbf{Mean F1} \\
        \hline
        \multirow{4}{*}{Source}
            & Good Dice \& Good XAI & 48 & 75.0 & 0.9055 & 0.7081 \\
            & Good Dice \& Bad XAI  &  6 &  9.4 & 0.7141 & 0.1147 \\
            & Bad Dice \& Good XAI  &  0 &  0.0 & --      & --      \\
            & Bad Dice \& Bad XAI   & 10 & 15.6 & 0.3927 & 0.1298 \\
        \hline
        \multirow{4}{*}{Target}
            & Good Dice \& Good XAI & 17 &  2.6 & 0.6463 & 0.4802 \\
            & Good Dice \& Bad XAI  &  9 &  1.4 & 0.6427 & 0.2435 \\
            & Bad Dice \& Good XAI  & 20 &  3.1 & 0.4924 & 0.4531 \\
            & Bad Dice \& Bad XAI   & 606 & \textbf{92.9} & 0.0681 & 0.0325 \\
        \hline
    \end{tabular}
\end{table}

\begin{figure*}[t]
\centering

\begin{subfigure}{\textwidth}
    \centering
    \includegraphics[width=\textwidth]{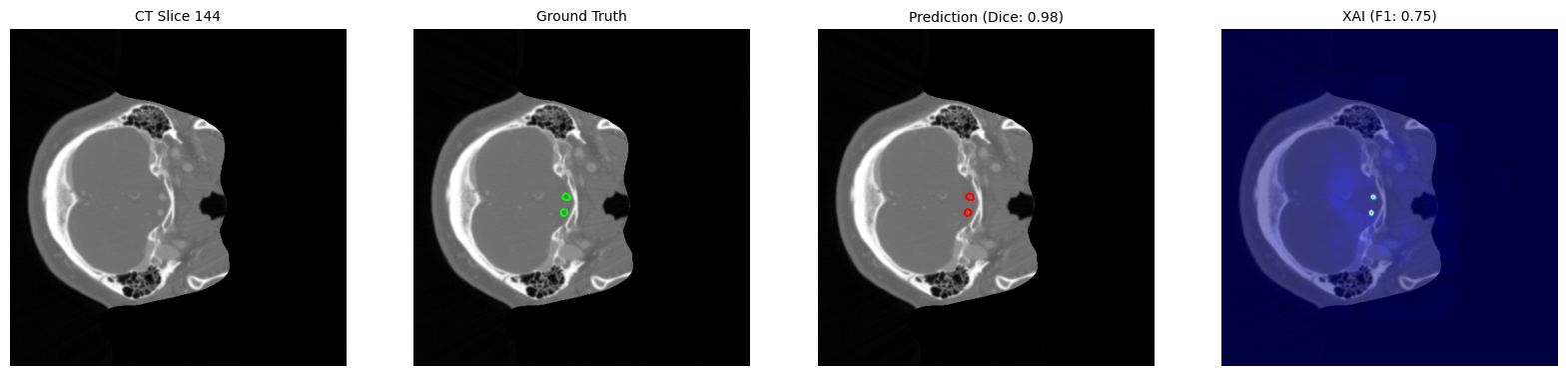}
    \caption{Good Dice \& Good XAI (75.0\%) — Attention correctly localizes on vessel structures, indicating proper feature reliance.}
    \label{fig:source_good_good}
\end{subfigure}

\vspace{0.2cm}

\begin{subfigure}{\textwidth}
    \centering
    \includegraphics[width=\textwidth]{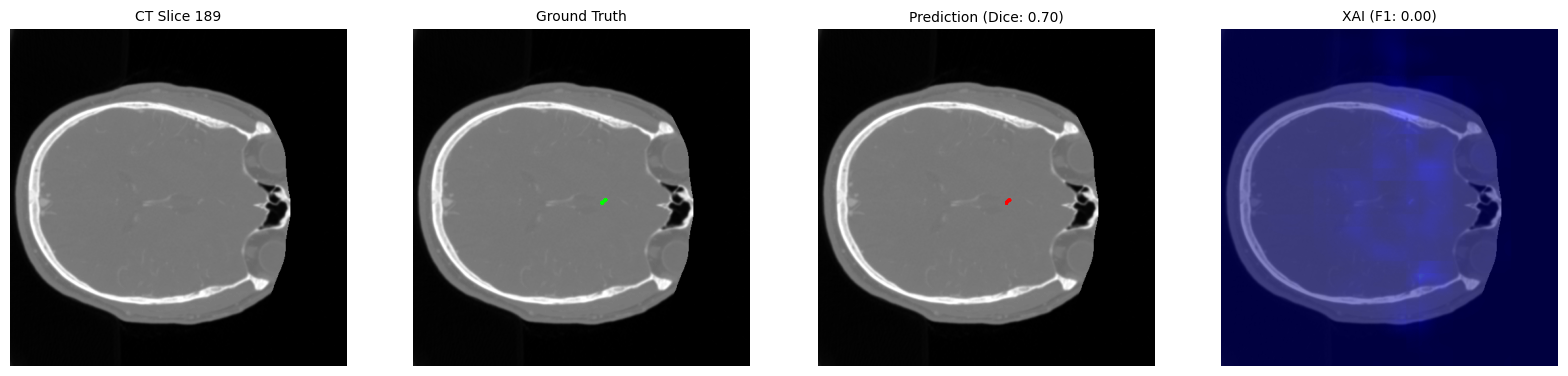}
    \caption{Good Dice \& Bad XAI (9.4\%) — Correct segmentation but diffuse, unfocused attention suggesting implicit feature learning.}
    \label{fig:source_good_bad}
\end{subfigure}

\vspace{0.2cm}

\begin{subfigure}{\textwidth}
    \centering
    \includegraphics[width=\textwidth]{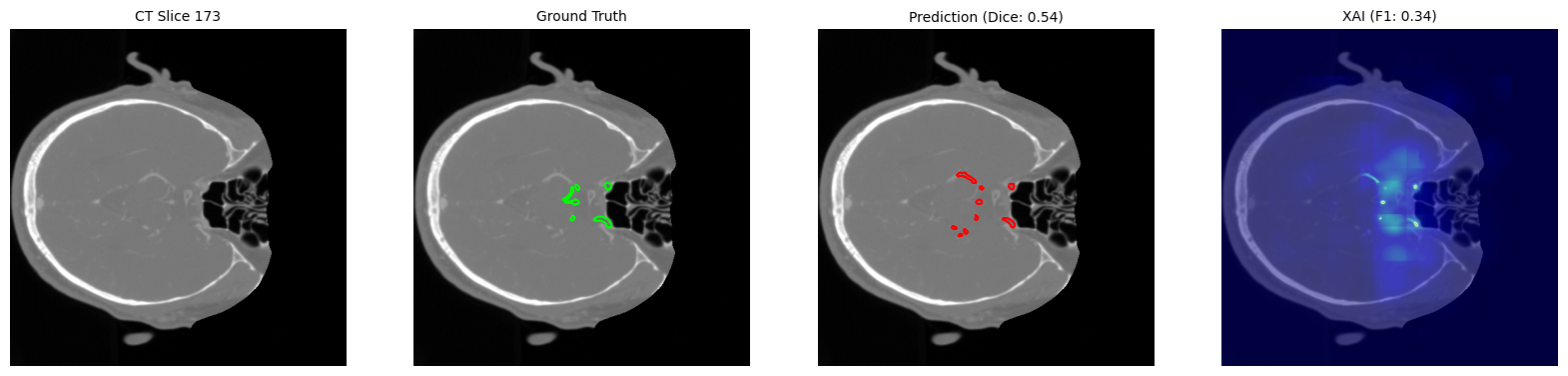}
    \caption{Bad Dice \& Bad XAI (15.6\%) — Both segmentation and attention fail, typically on challenging boundary cases.}
    \label{fig:source_bad_bad}
\end{subfigure}

\caption{Source domain (RSNA) Seg-XRes-CAM analysis across all four diagnostic categories. The majority of slices (75\%) exhibit both accurate segmentation and focused attention on vessel anatomy, confirming proper feature reliance on the training distribution.}
\label{fig:source_xai}
\end{figure*}

\begin{figure*}[t]
\centering

\begin{subfigure}{\textwidth}
    \centering
    \includegraphics[width=\textwidth]{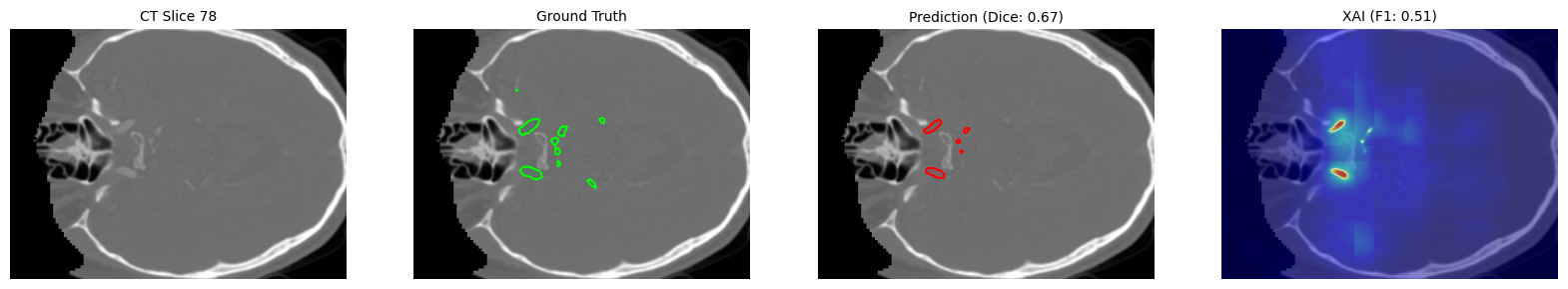}
    \caption{Good Dice \& Good XAI (2.6\%) — Rare cases where attention and segmentation both succeed on target data.}
    \label{fig:target_good_good}
\end{subfigure}

\vspace{0.2cm}

\begin{subfigure}{\textwidth}
    \centering
    \includegraphics[width=\textwidth]{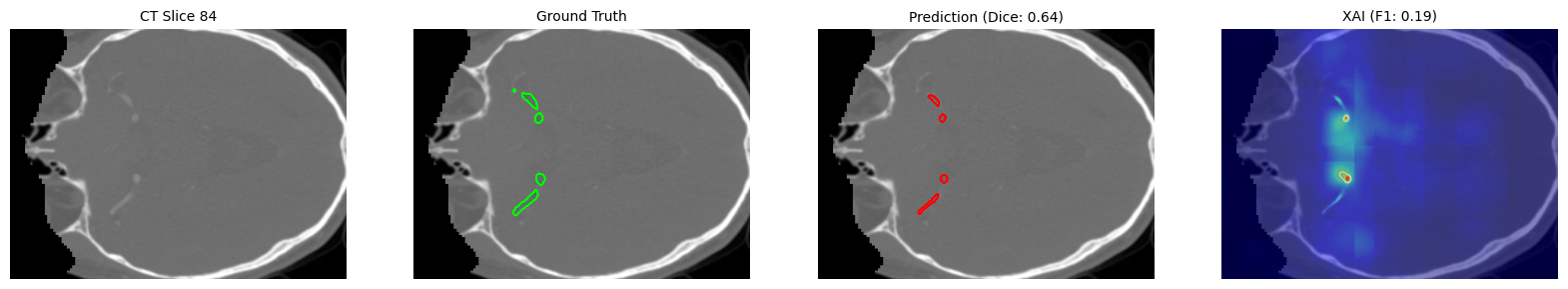}
    \caption{Good Dice \& Bad XAI (1.4\%) — Correct segmentation but attention fails to localize on vessel structures.}
    \label{fig:target_good_bad}
\end{subfigure}

\vspace{0.2cm}

\begin{subfigure}{\textwidth}
    \centering
    \includegraphics[width=\textwidth]{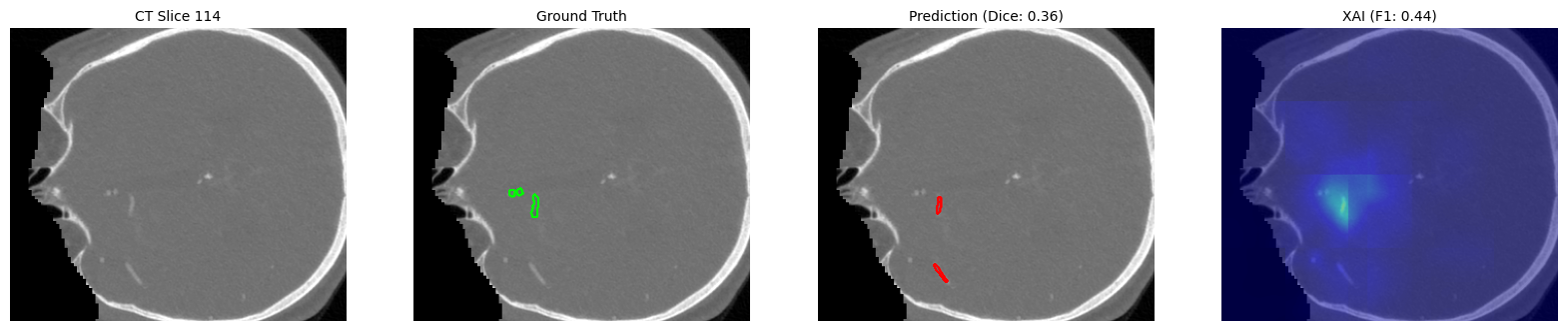}
    \caption{Bad Dice \& Good XAI (3.1\%) — Attention localizes correctly but segmentation still fails, indicating decoder-level issues.}
    \label{fig:target_bad_good}
\end{subfigure}

\vspace{0.2cm}

\begin{subfigure}{\textwidth}
    \centering
    \includegraphics[width=\textwidth]{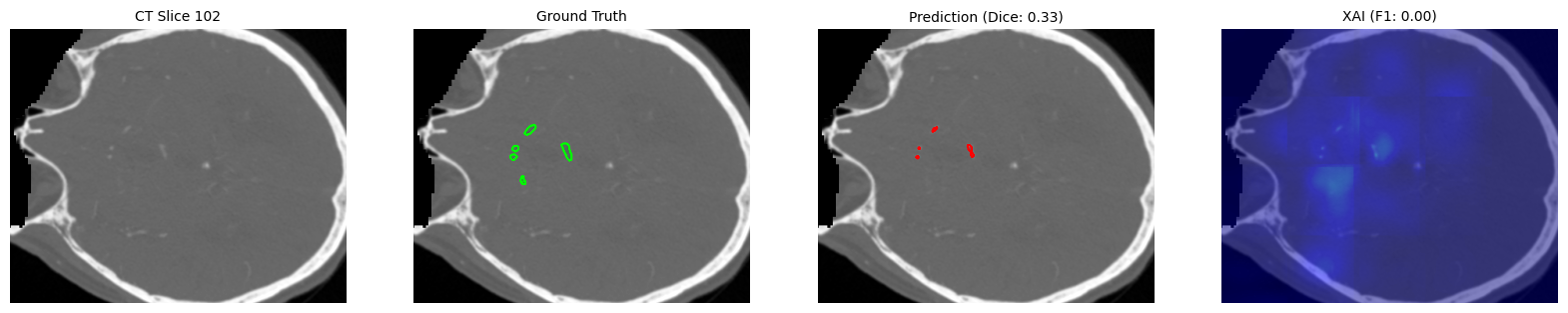}
    \caption{Bad Dice \& Bad XAI (92.9\%) — Systematic failure: attention shifts to anatomically irrelevant regions, confirming spurious correlation.}
    \label{fig:target_bad_bad}
\end{subfigure}

\caption{Target domain (TopCoW) Seg-XRes-CAM analysis across all four diagnostic categories. The overwhelming majority of slices (92.9\%) fall into the worst category, demonstrating that the model's attention has abandoned true anatomical features. This visual evidence confirms the quantitative finding: the model learned spurious, domain-specific correlations rather than generalizable vessel features.}
\label{fig:target_xai}
\end{figure*}

Representative qualitative examples (Figures~\ref{fig:source_xai} and \ref{fig:target_xai}) further support these results. On the Source domain, the attention maps correctly focus on vessel regions when the model produces accurate predictions. In contrast, on the Target domain, the attention frequently shifts to irrelevant structures, indicating that the learned features do not transfer across domains. These visual findings are consistent with the quantitative analysis and confirm the mechanisms driving the model’s generalization failure.


\section{Conclusion}
In this study, we addressed the critical problem of domain shift and generalization failure in high-performance State-Space Segmentation Models (SSMs) when transitioning from a complex, single-source dataset (RSNA CTA Aneurysm) to a clinically distinct external dataset (TopCoW Circle of Willis CT). This problem is highly significant because it directly compromises the trustworthiness and clinical applicability of modern deep learning solutions. The observed failure is not unique to cerebrovascular segmentation or SSMs; rather, it highlights the systemic tendency of advanced architectures to learn spurious correlations instead of robust, domain-invariant features. Our work, therefore, opens new research avenues focused on the diagnostic evaluation and architectural debiasing of emerging models across various medical specialties.

We executed a two-phase diagnostic approach: first, quantifying the domain gap and performance failure, and second, employing XAI to diagnose the root cause of the shift. The results of the first phase conclusively established a significant, quantifiable domain gap driven by three key factors: a 36\% reduction in Z-resolution, a 3.3× increase in background noise levels, and substantial differences in vessel diameter variability between the two datasets. This domain shift led to the catastrophic collapse of the model's performance, with the Dice score dropping from 0.8604 on the Source domain to 0.2902 on the Target domain.

The purpose of the second phase was to break the black-box nature of the State-Space model and determine why its decision-making process failed. This required developing a robust method for evaluating explainability maps to ensure the visualized attention was trustworthy, even on out-of-distribution data. The model under scrutiny was UMamaba, an example of an SSM, and the explainability method we employed was Seg-XRes-CAM with multi-layer mean aggregation, chosen for its relevance to segmentation and its capacity to smooth noisy gradients.

We evaluated our explainability maps by calculating the Intersection over Union (IoU) between the attention map and two masks: the Ground Truth mask (XAI-GT), and the model's Prediction Mask (XAI-PM). This provided a quantitative measure of whether the model's attention aligned with the correct anatomy or only with its own (often incorrect) output. Re-iterating the results of the second phase, the analysis provided visual and quantitative proof of the failure mechanism. On the Target domain, the model's attention was almost entirely unaligned with the true anatomical target, confirmed by the low XAI-GT IoU ($\approx0.101$). Conversely, the attention remained moderately aligned with its own erroneous predictions, evidenced by the higher XAI-PM IoU ($\approx0.282$). This significance lies in proving that the SSM overfit to dataset-specific spurious correlations (e.g., background noise and Z-artifacts) and abandoned true feature reliance when faced with a new distribution.

Moving forward, the diagnostic insights derived from this XAI analysis will be instrumental in guiding the development of conceptually sound and robust domain generalization strategies for emerging state-space architectures.


\bibliographystyle{unsrt}  
\bibliography{references}

\end{document}